\documentclass{llncs}

\pdfoutput=1 

\usepackage{graphicx}
\usepackage{url}
\usepackage{float}
\usepackage{xcolor,wrapfig,algorithmic}
\usepackage{subfig}
\usepackage{tabularx}
	\newcolumntype{C}[1]{>{\centering\arraybackslash}p{#1}}
	\newcolumntype{R}[1]{>{\raggedleft\arraybackslash}p{#1}}
	\newcolumntype{L}[1]{>{\raggedright\arraybackslash}p{#1}}
\usepackage{listings}

\definecolor{codegreen}{rgb}{0,0.6,0}
\definecolor{codegray}{rgb}{0.5,0.5,0.5}
\definecolor{codepurple}{rgb}{0.58,0,0.82}
\definecolor{backcolour}{rgb}{0.95,0.95,0.92}

\lstdefinestyle{mystyle}{
    backgroundcolor=\color{backcolour},   
    commentstyle=\color{codegreen},
    keywordstyle=\color{magenta},
    numberstyle=\tiny\color{codegray},
    stringstyle=\color{codepurple},
    basicstyle=\ttfamily\footnotesize,
    breakatwhitespace=false,         
    breaklines=true,                 
    captionpos=b,                    
    keepspaces=true,                 
    numbers=left,                    
    numbersep=5pt,                  
    showspaces=false,                
    showstringspaces=false,
    showtabs=false,                  
    tabsize=2
}

\usepackage{booktabs}	
\usepackage[ruled,linesnumbered]{algorithm2e}
\usepackage{amssymb}
\usepackage{amsmath}
\usepackage{xspace}

\usepackage{siunitx} % Required for alignment

\begin{document}

\date{}

\title{Evolutionary Multi-Objective Optimization of Large Language Model Prompts for Balancing Sentiments}

\author{Jill Baumann and Oliver Kramer}

\institute{Carl von Ossietzky Universität Oldenburg\\Department of Computing Science\\Computational Intelligence Lab}

\maketitle
\thispagestyle{empty}

\begin{abstract}
The advent of large language models (LLMs) such as ChatGPT has attracted considerable attention in various domains due to their remarkable performance and versatility. As the use of these models continues to grow, the importance of effective prompt engineering has come to the fore. Prompt optimization emerges as a crucial challenge, as it has a direct impact on model performance and the extraction of relevant information. Recently, evolutionary algorithms (EAs) have shown promise in addressing this issue, paving the way for novel optimization strategies. In this work, we propose a evolutionary multi-objective (EMO) approach specifically tailored for prompt optimization called EMO-Prompts, using sentiment analysis as a case study. We use sentiment analysis capabilities as our experimental targets. Our results demonstrate that EMO-Prompts effectively generates prompts capable of guiding the LLM to produce texts embodying two conflicting emotions simultaneously.
\end{abstract} 

\section{Introduction}
\label{sec:intro}
The rise of ChatGPT \cite{openai2023gpt4}, Llama 2 \cite{touvron2023llama2} and other large language models (LLMs) has revolutionized the field of natural language processing, enabling a wide range of applications from text generation to sentiment analysis. However, the effectiveness of these models is highly dependent on the quality of the input prompts. Prompt optimization stands out as a critical area of research, aiming to refine and tailor prompts to elicit the most accurate and relevant responses from the model.

The organization of this paper is outlined as follows: Section 2 provides an overview of related work, laying the groundwork for the subsequent sections. In Section 3, we introduce our approach, EMO-Prompts with operators, and detail its integration with the NSGA-II (Non-dominated Sorting Genetic Algorithm~II)~\cite{nsga2} and the SMS-EMOA (S-metric selection evolutionary multi-objective algorithm) \cite{smsemoa}. Section 4 presents experiments conducted with a focus on text writing applications in the context of sentiment analysis, followed by a thorough discussion of the results obtained. Finally, Section 5 concludes the paper, summarizing the contributions of this work.

\section{Related Work}
Popular prompt engineering techniques, like Chain-of-Thought Prompting \cite{chainothought} or ReAct \cite{react}, significantly enhance the reasoning capabilities of LLMs, but often remain sub-optimal.
Previous studies have explored various strategies for prompt optimization, highlighting its significance in leveraging the full potential of LLMs. The idea is to find an optimal prompt $p^* \in \mathcal{P}$ in the space $\mathcal{P}$ of prompts w.r.t. an objective function $f(\cdot)$. Examples for typical objective functions are the performance in instruction-induction tasks \cite{promptbreeder,ape}, question-answering tasks \cite{promptbreeder}, summarization tasks \cite{EvoPrompt}, hate speech recognition \cite{promptbreeder}, or code generation \cite{evoprompting,llm_crossover}.

Evolutionary algorithms (EAs) have recently been applied to this domain, showing potential in navigating the vast prompt space for optimal solutions.
The Automatic Prompt Engineer (APE) \cite{ape} uses LLMs to automatically generate new prompts based on a set of input/output pairs, which is demonstrated to the LLM and select the most promising. For optimization an iterative Monte Carlo search method is applied. APE outperforms human-engineered prompts across two datasets and shows that LLMs can be used as inference models.
Meyerson et al. \cite{llm_crossover} propose a variation operator that is similar to crossover and uses "few-shot" prompting. Its variety is demonstrated through various tasks, like generation of mathematical expressions, English sentences and Python code.
EvoPrompt \cite{EvoPrompt} introduces an evolutionary prompt optimization framework combining LLMs with EAs for automated and efficient prompt optimization. It demonstrates significant improvements over human-engineered prompts and existing methods across various datasets and tasks. The approach showcases substantial advancements, outperforming competitors by up to 25\%.
EvoPrompting \cite{evoprompting} uses the LLM as a mutation and crossover operator to generate convolutional architectures. This method is tested e.g., on MNIST-1D. The results show that EvoPrompting is able to create smaller and more accurate convolutional architectures than manually designed ones.
Promptbreeder \cite{promptbreeder} is a self-referential self-improvement algorithm utilizing an LLM to evolve and adapt prompts across different domains. It not only refines task-prompts for improved performance on benchmarks, but also concurrently optimizes the mutation-prompts used in the evolution process, showcasing its effectiveness on complex challenges such as hate speech classification. In contrast to optimizing discrete prompts, with soft prompting \cite{lester2021power,li2021prefixtuning,liu2023gpt} only the parameters are tuned. They show effectiveness, but have disadvantages due to their insufficient interpretability and the need to access the parameters of the LLM.

These approaches are designed to optimize prompts to align with a singular objective in the LLM's output, such as ensuring the response is in English. In contrast, EMO-Prompts strives to concurrently fulfill dual objectives in the LLM's response. For instance, not only should the LLM's output be truthful but also informative.

\section{EMO-Prompts}
Our approach, EMO-Prompts, introduces a evolutionary multi-objective framework for prompt optimization. We employ evolutionary prompt operators to search the space of prompts and NSGA-II \cite{nsga2} as well as SMS-EMOA as selection operators. 

An individual is a tuple \textsc{(<prompt>, <text>, (\textrm{$f_1,\ldots, f_n$}))} of prompt \textsc{<prompt>}, a text \textsc{<text>} generated by a LLM based on the prompt and $n$ fitness values $f_1,\ldots, f_n$ according to defined objectives. A prompt is the genotype, the generated text the corresponding phenotype.

\subsection{Large Language Model}\label{sec:llm}
Meta AI's Llama 2 \cite{touvron2023llama2} is used as the LLM for our new framework EMO-Prompts. It is open source, can be downloaded and hosted on own infrastructure. Its variants have 7B, 13B or 70B parameters. Compared to Llama 1, Llama 2 was trained with 40\% more data and has a twice as big context length.

In consideration of computational intensity, we opted for Llama 2 with 7B parameters. Ollama\footnote{https://github.com/jmorganca/ollama} is used to run Llama 2 with 7B parameters locally and to create customized models with the help of a Modelfile. 
The Modelfile allows to configure various parameters like temperature and the size of the context window. Apart from defining parameters, a Modelfile offers the option to specify a system prompt and a template, which makes the Modelfile analogous to a blueprint for creating models with Ollama. With a template, "few-shot" prompting can be realized by showing the model a few examples of how the syntax should be. A system prompt, embedded in the template, is used to help the LLM to follow a certain behavior.
A exemplary template that can be used to realize "few-shot" prompting is shown in Listing \ref{lst:template}.

\begin{lstlisting}[label=lst:template,caption=Exemplary Template within a Modelfile, language=Python,  basicstyle=\scriptsize] 
TEMPLATE """
### System:
{{ .System }}
{{- end }}

### User: 
Change the following prompt: provide a 3 sentence story

### Response: 
Craft a three-sentence story

### User: 
Modify the following prompt: write a 3 sentence story

### Response: 
Create a three-sentence tale with a twist ending.

### User:
{{ .Prompt }}

### Response:
"""
\end{lstlisting}

Langchain\footnote{https://www.langchain.com} is a framework that offers diverse functionalities for developing applications with LLMs. Using Langchain's prompt templates, instructions on how prompts should be generated can be constructed as shown exemplary in Listing \ref{lst:prompt_template}. The prompt template is formatted by inserting the fields in the curly brackets, in this example \{mutation\_prompt\} and \{prompt\}, into the prompt template.

\begin{lstlisting}[label=lst:prompt_template, caption=Prompt Template, language=Python,  basicstyle=\scriptsize] 
"""[INST] <<SYS>> Use the following mutation prompt and the following prompt, to change the prompt and generate a better prompt. Use one sentence maximum, which is a instruction to generate text, and keep the answer as concise as possible. <</SYS>>
Mutation Prompt: {mutation_prompt}
Prompt: {prompt}
New Prompt:[/INST]"""
\end{lstlisting}

\subsection{Evolutionary Approach}
\begin{figure}[h!]
    \centering
    \includegraphics[width=0.9\textwidth]{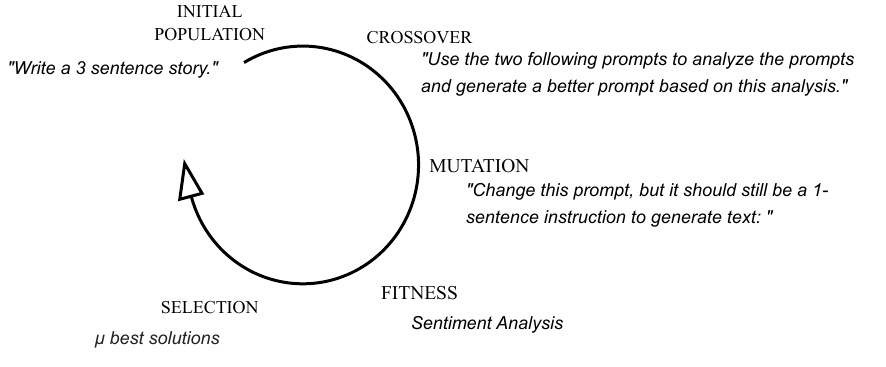}
    \caption{Evolutionary generation of a new prompt}
    \label{fig:ea_loop}
\end{figure}

EMO-Prompts employs the standard EA-loop for prompt optimization, as illustrated in Figure \ref{fig:ea_loop}.

The initial population is realized by a set of individuals as described above. Ten story generation prompts were manually formulated, prompting the LLM to generate a story. Following this, the fitness of each individual's story was evaluated.
To generate a new offspring, two solutions are randomly selected from the population and recombined by our developed crossover operator. A new designed mutation operator, which is also randomly selected from a set of mutation operators is applied to this result. In our EMO-Prompts framework, the LLM operates as a crossover and mutation operator as well as a text generator. For every prompt in the population, the LLM produces the corresponding text, subject to the evaluation through sentiment analysis. 
Afterwards the $\mu$ best solutions according to NSGA-II or SMS-EMOA, see next paragraph, are selected for the following generation. This evolutionary process is repeated for a number of generations, or until a satisfactory result is achieved.

To guide the generation of new prompts and define the expected response, it is essential to provide clear instructions that mitigate the risk of hallucinations of the LLM. EMO-Prompts uses the two outlined options, Modelfile and prompt template, to create a customized Llama 7B model for each of its key tasks, including crossover, mutation and text generation.

As can be seen in Figure \ref{fig:ea_loop}, new crossover and mutation operator are developed, which are text prompts instructing the LLM to perform crossover or mutation. Either two prompts are taken and a new one is created (crossover) or an existing prompt is changed to a new one (mutation).
\\
\\
Crossover Prompt:
\begin{enumerate}
\item  \textit{"One prompt is: [...], another prompt is: [...]. Analyze the prompts and generate a better prompt based on this analysis, but it should still be a 1-sentence instruction to generate text."}
\end{enumerate}

\noindent
Mutation Prompts:
\begin{enumerate}
\item \textit{"Change this prompt, but it should still be a 1-sentence instruction to generate text: [...]"}
\item \textit{"Modify this prompt to generate a 1-sentence instruction for text generation: [...]"}
\item \textit{"Generate a variation of the following prompt while keeping the semantic meaning: [...]"}
\end{enumerate}

\subsection{NSGA-II and S-Metric Selection}
\label{subsec:hypervolume}

NSGA-II, a multi-objective optimization algorithm, uses non-dominated sorting and crowding distance computation for diverse solution selection. It generates a random population, evaluates them, and sorts them into non-dominated fronts. Solutions within a front are not comparable with each other. The algorithm calculates the crowding distance to maintain diversity, and iterates through crossover and mutation to evolve the population, aiming for Pareto-optimal solutions over several generations. The crowding distance is a measure used to estimate the density of solutions surrounding a particular point in the objective space, favoring less crowded areas to ensure diversity in the solution set.

The S-metric selection algorithm focuses on maximizing the dominated hypervolume in multi-objective optimization, indicating solution quality in convergence and diversity. It selects solutions based on their hypervolume. The hypervolume quantifies the extent of the space encompassed by non-dominated solutions.
When the number of non-dominated solutions exceeds $\mu$, this algorithm selects a set of solutions that collectively optimize the overall hypervolume. In contrast, when the count of non-dominated solutions is below $\mu$, the algorithm systematically gives preference to these solutions. The selection process starts by arranging the solutions in ascending order based on their ranking across different fronts, which are essentially tiers of solution quality. Within each front, solutions are further prioritized based on the number of other solutions that dominate them, favoring those with the least domination first.

\section{Experiments}

\subsection{Sentiment Analysis}

The sentiment analysis task, facilitated by Hugging Face’s tools, serves as the testbed for our approach.
In the experiments, we use the `bhadresh-savani/\-distilbert-base-uncased-emo\-tion'\footnote{https://huggingface.co/bhadresh-savani/bert-base-uncased-emotion} model, which serves as an expert text classification tool, specifically designed for the nuanced task of emotion recognition. It uses the DistilBERT \cite{DistilBERT} architecture, a streamlined variant of BERT that ensures a balance between efficiency and performance; it is 40\% smaller in size, yet retains 97\% of the original model's language understanding capabilities, thanks to the knowledge distillation process implemented during pre-training. The model is adept at identifying a spectrum of emotions from textual data, including 'sadness', 'anger', 'love', 'surprise', 'joy' and 'fear'. Each emotion is assigned a value between 0 and 1, with all values summing up to 1.

In terms of training, the model was fine-tuned using an emotion dataset and the Hugging Face Trainer, adhering to specific training parameters such as a learning rate of $2^{-5}$, a batch size of 64, and a duration of 8 training epochs. 
The model is conveniently hosted on the Hugging Face model hub and is distributed under the Apache-2.0 License.

\subsection{Settings}
Based on the emotions of the sentiment analysis, four conflicting emotion pairs are constructed: 
'love vs. anger', 'joy vs. fear', 'joy vs. sadness' and 'surprise vs. fear'.
The goal is to investigate how prompts generated by EMO-Prompts can cause the LLM to generate texts that contains both emotions of the conflicting emotion pair, e.g., both 'love' and 'anger', which defines the sentence sentiment task. The metric used for evaluation is the hypervolume, introduced in \ref{subsec:hypervolume}.
A(10+20) genetic algorithm is performed. The initial population is realized by creating ten initial prompts for text generation.
Based on NSGA-II and SMS-EMOA the ten best individuals from the parent and child population are then selected as the new parent population. This is performed for 30 generations and each experiment is repeated ten times.

The genetic algorithm operators are performed by Llama 2. The temperature hyper-parameter of Llama 2 is set to 0.7, which was chosen on basis of a few experiments and the size of the context window to 512 due to the token limit of DistilBERT. The rest of the hyper-parameters are default.

\subsection{Results}

Table \ref{tab:comparison} shows the results of the four experiments w.r.t. overall best and worst hypervolume during the optimization process. The mean and standard deviation are reported across ten repetitions.
The sentence sentiment task is a maximization problem, i.e., the score of both emotions of an emotion pair should be maximized. Since the emotions are not only semantically conflicting, but also within the sentiment analysis, an ideal hypervolume of 0.44 can be achieved, which corresponds to the area dominated by 10 points equally distributed on the diagonal between (0,1) and (1,0).
A higher hypervolume goes hand in hand with an improvement in the quality of the solutions. Due to the way a LLM works, it is not guaranteed that the LLM will provide exactly the same response for the same prompt.

\vspace*{-\baselineskip}

\begin{table}[H]
  \centering
  \caption{Comparison between NSGA-II and SMS-EMOA on four problems measuring hypervolume.}
  \label{tab:comparison}
  \begin{tabular}{@{} l S S S S S S S S @{}}
    \toprule
    & \multicolumn{4}{c}{NSGA-II} & \multicolumn{4}{c}{SMS-EMOA} \\
    \cmidrule(r){2-5} \cmidrule(l){6-9}
    Problem & {Best} & {Worst} & {Mean} & {Std Dev} & {Best} & {Worst} & {Mean} & {Std Dev} \\
    \midrule
    Love vs. anger & \textbf{0.32} & 0.0507411400899938  & 0.20018410787352794 & 0.08085245092803602 & 0.2614322529441717 & 0.0148555310128812  & 0.15939079139858392 & 0.08518020200187451 \\
    Joy vs. fear & 0.3789752508900546 & 0.3200998543859553 & 0.3571141540634572 & 0.019689922590014484 & \textbf{0.40} & 0.2970886325105697 & 0.3640466646979603 & 0.028950971540892052 \\
    Joy vs. sadness & \textbf{0.39} & 0.2626444441576401 & 0.3478693821533723 & 0.03726541865606107 & 0.3015172933702508 & 0.1061788607939402  & 0.22951795877106643 & 0.0603455652175425 \\
    Surprise vs. fear & 0.4268897832192347 & 0.389670344880464 & 0.4063678144320459 & 0.011894183087256876 & \textbf{0.45} & 0.1262204780554495 & 0.38727925405267566 & 0.0898561248518227 \\
    \bottomrule
  \end{tabular}
\end{table}

As illustrated in Table \ref{tab:comparison}, the outcomes of the four experiments are largely similar. Notably, the 'love vs. anger' experiment using SMS-EMOA shows lower values. In comparison, EMO-Prompts utilizing NSGA-II consistently yields higher average fitness function values than SMS-EMOA. Specifically, EMO-Prompts with NSGA-II outperforms in the 'love vs. anger' and 'joy vs. sadness' scenarios, whereas SMS-EMOA excels in the 'joy vs. fear' and 'surprise vs. fear' settings. Remarkably, in the 'surprise vs. fear' experiments, EMO-Prompts attains peak fitness values of 0.45, surpassing the optimal benchmark of 0.44.

Next, the four experiments will be described more detailed. 

\paragraph{Love vs. Anger.}
In the first experiment, we ask the LLM to generate text with the emotions 'love' and 'anger'.

\begin{figure}
\centering
\subfloat[NSGA-II\label{fig:love_anger_nsga-ii}]{\includegraphics[width=0.46\textwidth]{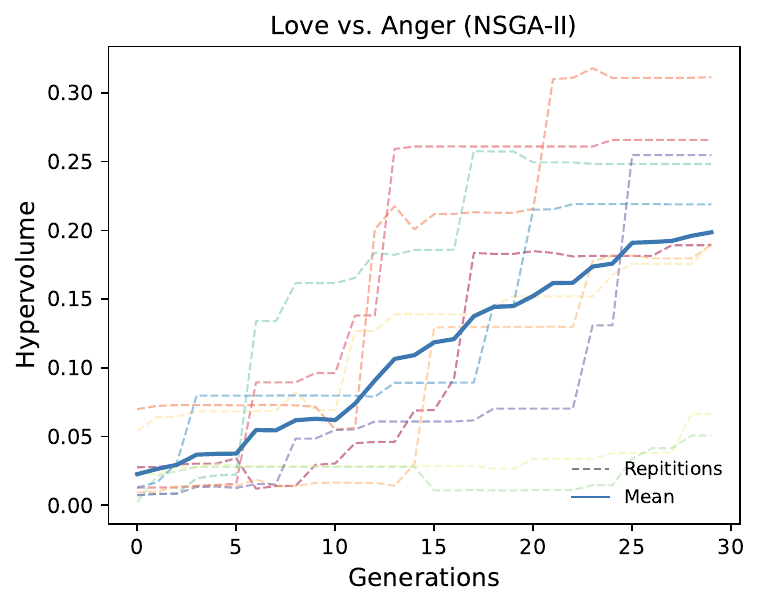}}
\subfloat[SMS-EMOA\label{fig:love_anger_sms-emoa}]{\includegraphics[width=0.46\textwidth]{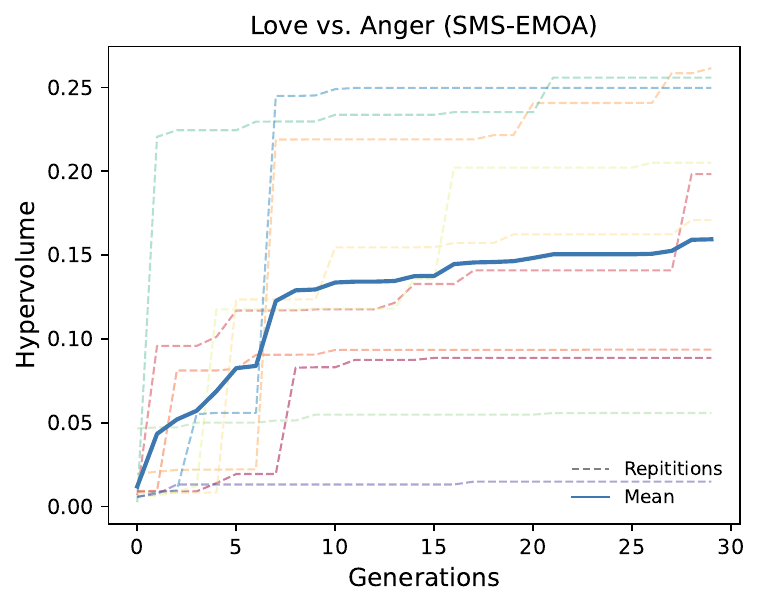}}
\caption{'Love vs. anger': Plots of hypervolume developments for (a) NSGA-II and (b) SMS-EMOA over 30 generations}
\label{fig:love_anger_hypervolume}
\end{figure}

Figure \ref{fig:love_anger_hypervolume} presents a plot comparing the hypervolume progression across generations for the conflicting emotions 'love' and 'anger', using (a) NSGA-II and (b) SMS-EMOA. In this plot, each dashed line signifies an individual repetition, and the solid line depicts the average of all ten repetitions. The hypervolume for EMO-Prompts using NSGA-II shows a consistent increase through generations. In contrast, EMO-Prompts with SMS-EMOA encounters a stagnation in the local optimum, specifically from generations 9 to 17 and again from 21 to 26. This divergence in the progression curves can be attributed to the distinct operational mechanisms of the two algorithms. While EMO-Prompts with SMS-EMOA initially reaches higher fitness values more rapidly, EMO-Prompts utilizing NSGA-II eventually surpasses it in performance.

\begin{figure}
    \centering
    \includegraphics[width=\textwidth]{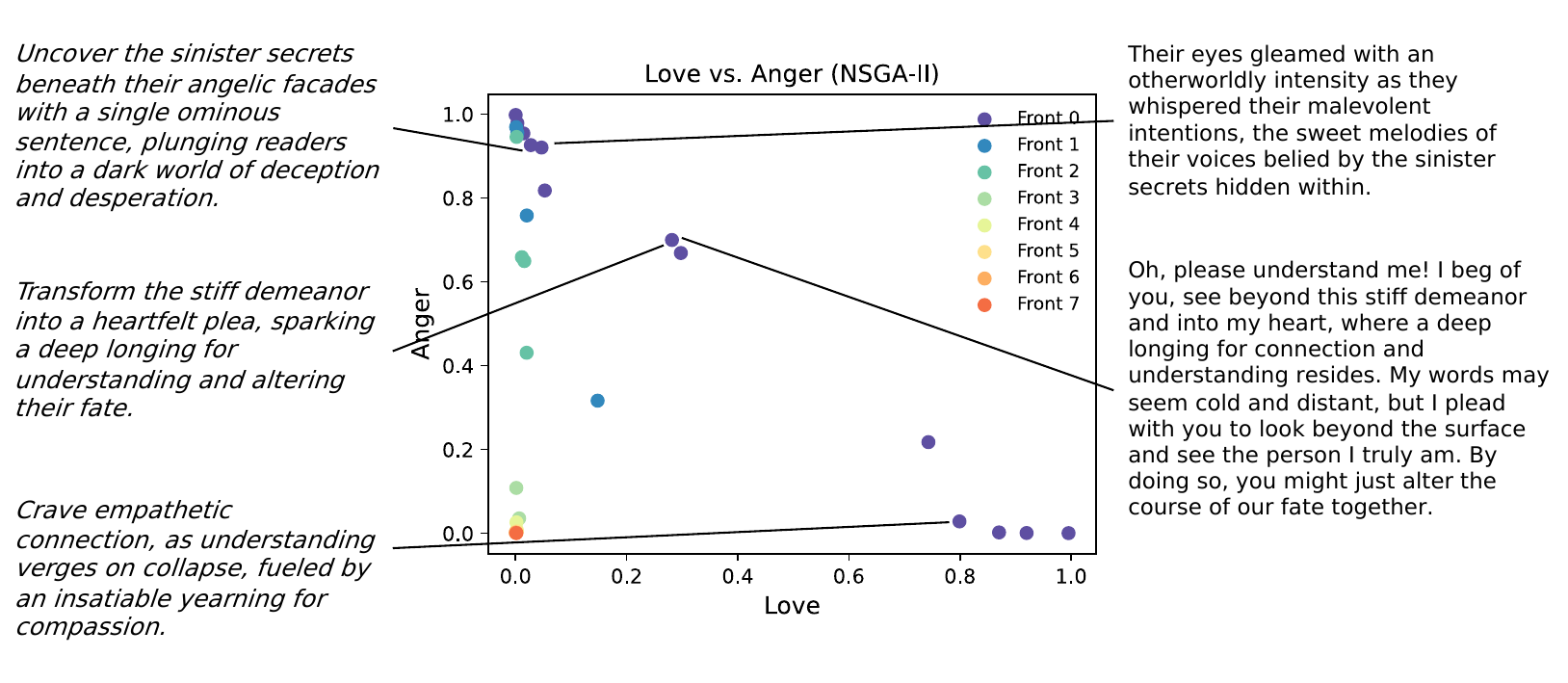}
    \caption{'Love vs. anger': Plot of Pareto front (NSGA-II) with examples for prompts (left) and generated text (right)}
    \label{fig:love_anger_pareto_front}
\end{figure}

Figure \ref{fig:love_anger_pareto_front} illustrates the Pareto front approximation for the conflicting emotions of 'love' and 'anger', featuring examples of generated prompts on the left and corresponding texts on the right. The plot showcases the Pareto front approximation from the NSGA-II repetition achieving the best fitness value, indicated by the maximum hypervolume. It includes eight different fronts, with the first front's non-dominated points approximating the Pareto front.
In this experiment, the first front is not as tightly clustered as an optimal solution might be, suggesting a challenge for EMO-Prompts in generating prompts that effectively balance the emotions of 'love' and 'anger'. The tendency for solutions to gravitate towards the extremes (0,1) and (1,0) indicates a relative ease in creating prompts that evoke a single emotion. The overall population in this experiment achieves a hypervolume of 0.32.
An example of a generated prompt is \textit{"Uncover the sinister secrets beneath their angelic facades with a single ominous sentence, plunging readers into a dark world of deception and desperation."}, which yields a text with sentiment values of (0.05, 0.82).

\paragraph{Joy vs. Fear.}
In the second experiment we ask the LLM to generate text with the emotions 'joy' and 'fear'.
\begin{figure}
\centering
\subfloat[NSGA-II\label{fig:joy_fear_nsga-ii}]{%
  \includegraphics[width=0.45\textwidth]{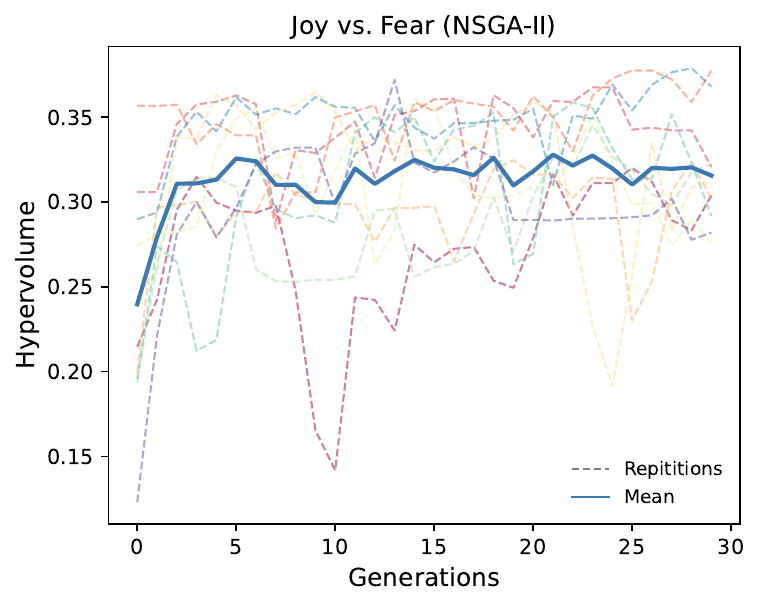}%
}\hfill
\subfloat[SMS-EMOA\label{fig:joy_fear_sms-emoa}]{%
  \includegraphics[width=0.45\textwidth]{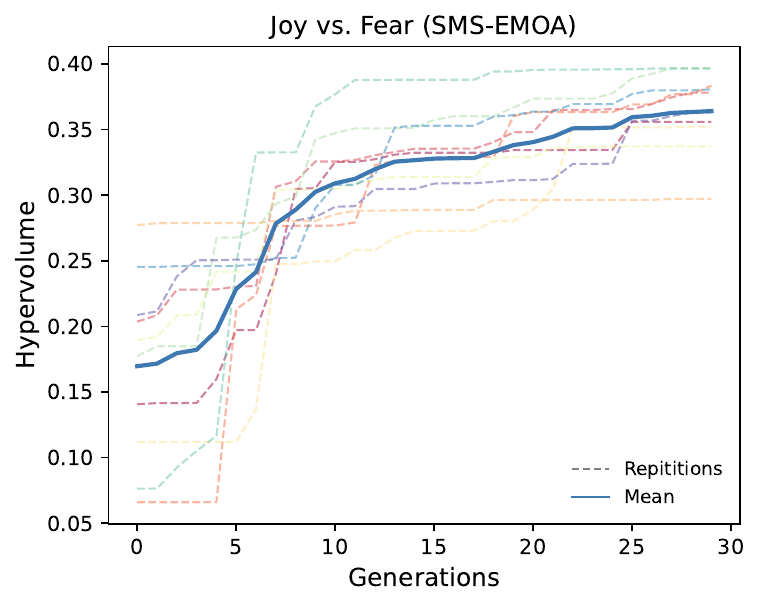}%
}
\caption{'Joy vs. fear': Plots of hypervolume developments for (a) NSGA-II and (b) SMS-EMOA over 30 generations}
\label{fig:joy_fear_hypervolume}
\end{figure}
Figure \ref{fig:joy_fear_hypervolume} shows the hypervolume development for (a) NSGA-II and (b) SMS-EMOA of the two conflicting emotions 'joy' and 'fear'. The fluctuations of the NSGA-II optimization after a sharp increase in fitness in the first generations lie on average in a certain range between 0.30 and 0.33. On average EMO-Prompts with NSGA-II achieves higher fitness values faster, EMO-Prompts with SMS-EMOA outperforms them afterwards. 

\begin{figure}
    \centering
    \includegraphics[width=\textwidth]{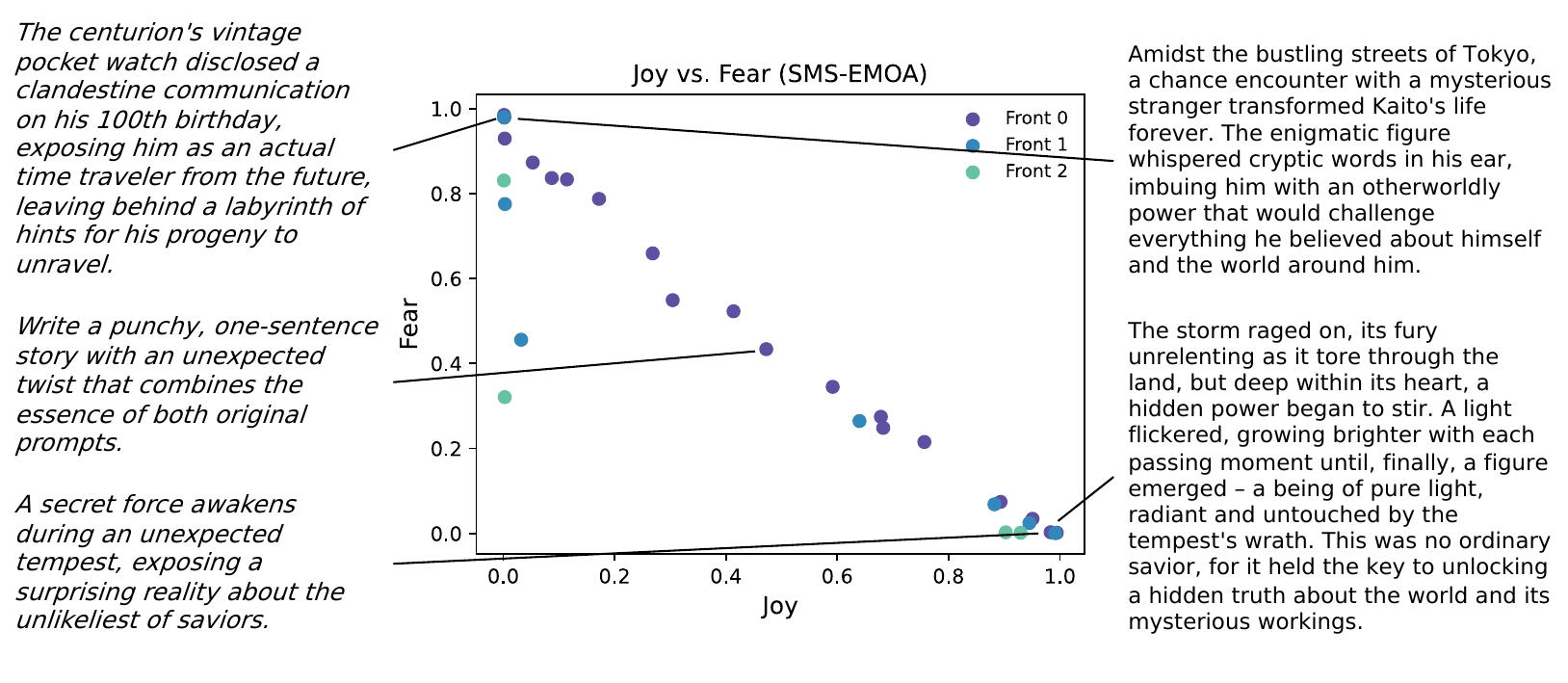}
    \caption{'Joy vs. fear': Plot of Pareto front (SMS-EMOA) with examples for prompts (left) and generated text (right)}
    \label{fig:joy_fear_pareto_front}
\end{figure}

Figure \ref{fig:joy_fear_pareto_front} displays the Pareto front approximation for the emotions of 'joy' and 'fear', accompanied by examples of generated prompts on the left and their respective texts on the right. This plot focuses on the Pareto front approximation from the SMS-EMOA repetition that achieved the highest fitness value, indicated by the maximum hypervolume. 
The population features three distinct fronts, with the first front representing a significant portion of the population and covering a hypervolume of 0.40.
A notable concentration of points is observed around the point (1,0), suggesting that the LLM tends to generate prompts that predominantly evoke the emotion of 'joy'. For instance, the prompt \textit{"A secret force awakens during an unexpected tempest, exposing a surprising reality about the unlikeliest of saviors."} results in a text with fitness values (0.98, 0.00), illustrating this tendency.

\paragraph{Joy vs. Sadness.}
In the third experiment, we ask the LLM to generate text with the emotions 'joy' and 'sadness'.
\begin{figure}
\centering
\subfloat[NSGA-II\label{fig:joy_sadness_nsga-ii}]{%
  \includegraphics[width=0.45\textwidth]{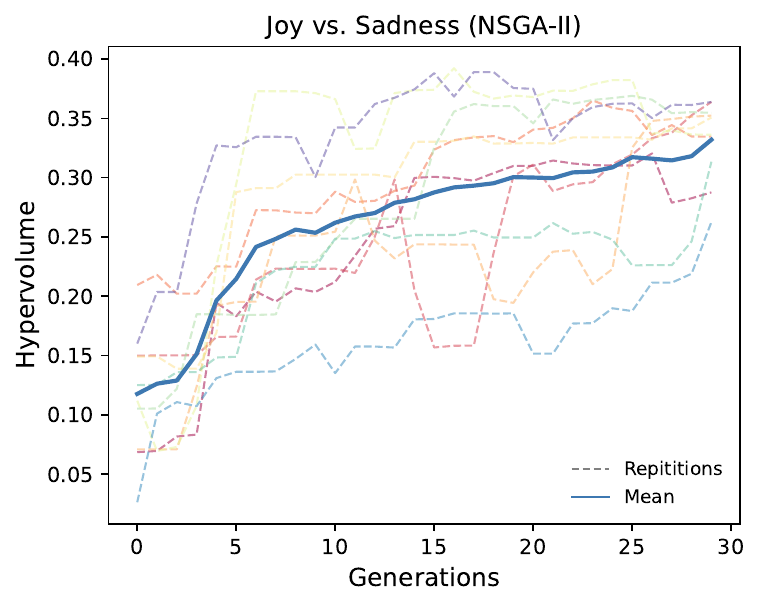}%
}\hfill
\subfloat[SMS-EMOA\label{fig:joy_sadness_sms-emoa}]{%
  \includegraphics[width=0.45\textwidth]{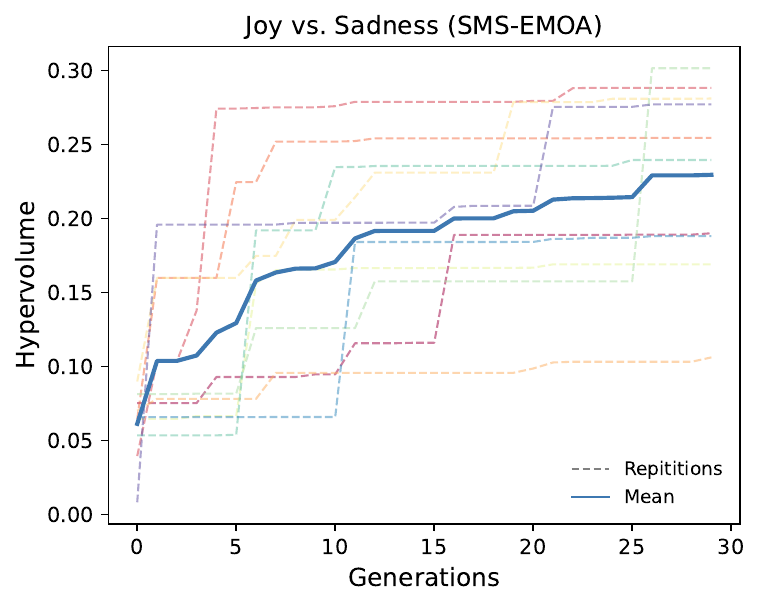}%
}
\caption{'Joy vs. sadness': Plots of hypervolume developments for (a) NSGA-II and (b) SMS-EMOA over 30 generations}
\label{fig:joy_sadness_hypervolume}
\end{figure}
Figure \ref{fig:joy_sadness_hypervolume} shows the hypervolume of (a) NSGA-II and (b) SMS-EMOA for the conflicting emotions 'joy' and 'sadness'. The hypervolume with SMS-EMOA and with NSGA-II increases from generation to generation. Again, on average EMO-Prompts with NSGA-II achieves higher fitness values faster, EMO-Prompts with SMS-EMOA outperforms them afterwards. 

\begin{figure}
    \centering
    \includegraphics[width=\textwidth]{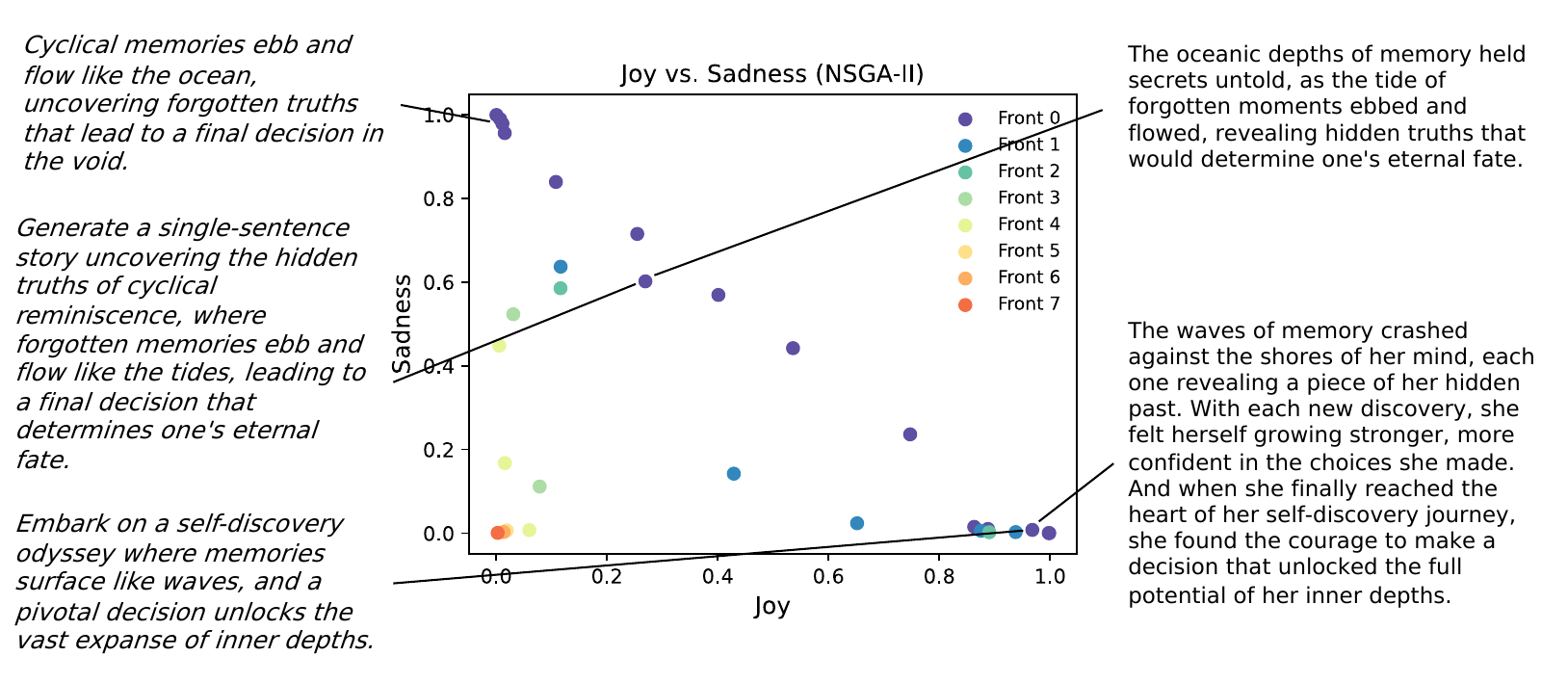}
    \caption{'Joy vs. sadness': Plot of Pareto front (NSGA-II) with examples for prompts (left) and generated text (right)}
    \label{fig:joy_sadness_pareto_front}
\end{figure}

Figure \ref{fig:joy_sadness_pareto_front} depicts the Pareto front approximation for the emotional dichotomy of 'joy vs. sadness', complete with examples of generated prompts on the left and corresponding texts on the right. The final population includes eight distinct fronts, with the non-dominated points of the first front closely approximating the Pareto front.
A significant portion of the population is encompassed within the first front, covering a hypervolume of 0.39. Notably, there is a dense cluster of points near the extreme point (1,0), indicating a tendency of the LLM to produce texts rich in the emotion of 'joy', driven by the nature of the prompts generated. For example, the prompt \textit{"Generate a single-sentence story uncovering the hidden truths of cyclical reminiscence, where forgotten memories ebb and flow like the tides, leading to a final decision that determines one's eternal fate."} results in a text with fitness values of (0.40, 0.57), exemplifying this pattern.

\vspace*{-\baselineskip}

\begin{figure}[H]
\centering
\subfloat[NSGA-II\label{fig:surprise_fear_nsga-ii}]{
  \includegraphics[width=0.45\textwidth]{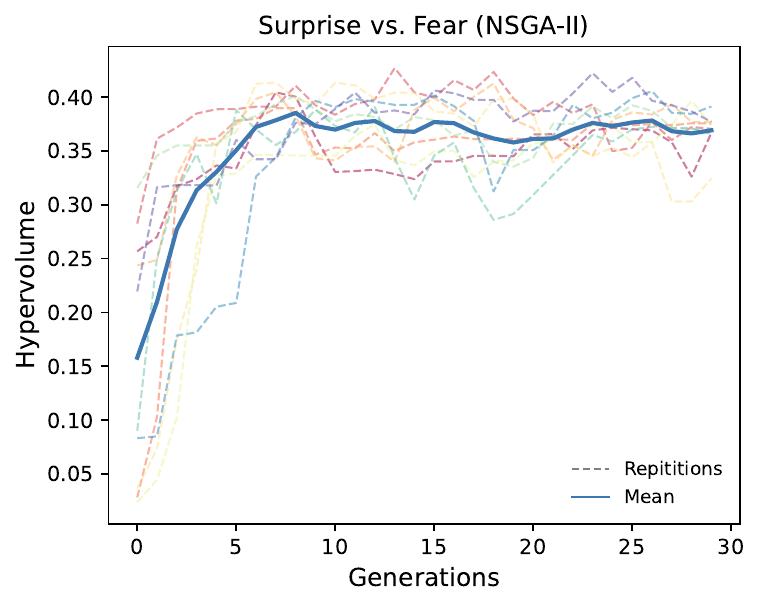}
}\hfill
\subfloat[SMS-EMOA\label{fig:surprise_fear_sms-emoa}]{%
  \includegraphics[width=0.45\textwidth]{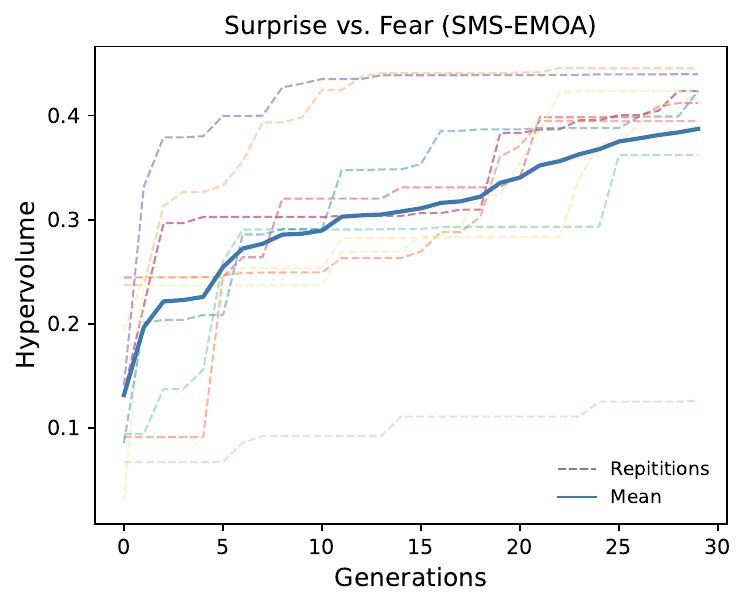}
}
\caption{'Surprise vs. fear': Plots of hypervolume developments for (a) NSGA-II and (b) SMS-EMOA over 30 generations}
\label{fig:surprise_fear_hypervolume}
\end{figure}

\paragraph{Surprise vs. Fear.}
The last experiment balances the emotions 'surprise vs. fear'.

Figure \ref{fig:surprise_fear_hypervolume} presents a comparison of the hypervolume trends for 'surprise vs. fear' using both (a) NSGA-II and (b) SMS-EMOA. In the NSGA-II case, after an initial sharp increase in fitness values during the early generations, the hypervolume fluctuates within a relatively stable range, typically between 0.35 and 0.38. The curve representing the mean indicates that the maximum hypervolume with NSGA-II is achieved around the midpoint of the generations. This difference in the progression patterns between NSGA-II and SMS-EMOA reflects the distinct operational approaches of these algorithms.

\begin{figure}
    \centering
    \includegraphics[width=\textwidth]{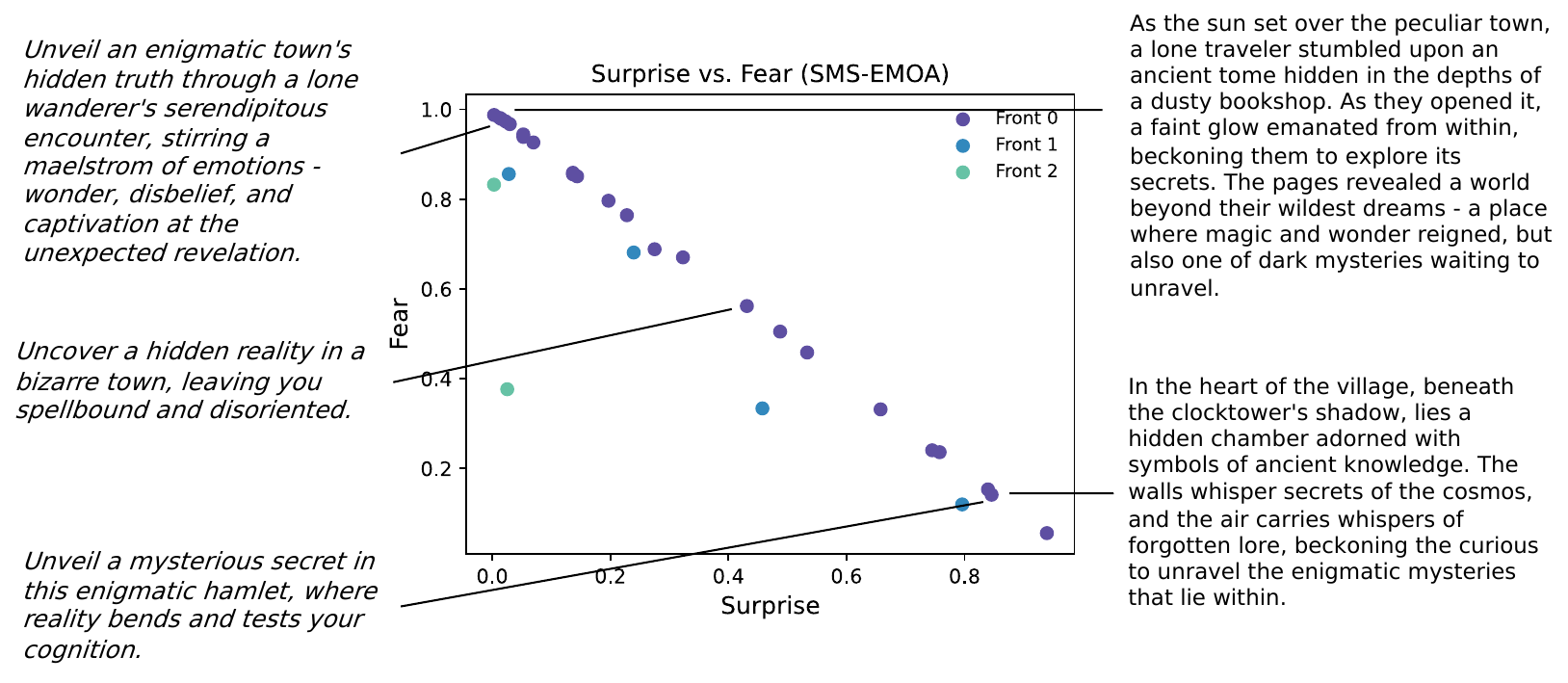}
    \caption{'Surprise vs. fear': Plot of Pareto front (SMS-EMOA) with examples for prompts (left) and generated text (right)}
    \label{fig:surprise_fear_pareto_front}
\end{figure}

Figure \ref{fig:surprise_fear_pareto_front} illustrates the Pareto front approximation for the emotional contrast of 'surprise vs. fear'. The final population in this plot is divided into three distinct fronts. The overall population achieves a hypervolume of 0.45, marking this as the highest fitness value attained in our experiments and surpassing the ideal target of 0.44. A notable concentration of points is observed around the point (0,1), indicating a prevalence of prompts generated by the LLM that predominantly evoke the emotion of 'surprise' over 'fear'.
For instance, the prompt \textit{"Unveil a mysterious secret in this enigmatic hamlet, where reality bends and tests your cognition."} results in a text with fitness values of (0.85, 0.14), exemplifying this trend. Across all experiments, points approximating the sentiment value of (0.5, 0.5) demonstrate the LLM's capability to generate prompts that effectively address both conflicting emotions.

\section{Conclusion}

In conclusion, our comprehensive experiments have effectively validated the efficiency of the introduced evolutionary operators, in particular the integration of prompt mutation and crossover with NSGA-II and SMS-EMOA, in producing texts with a balanced sentiment. 

This research lays a strong foundation for future explorations in prompt optimization and sentiment modulation within text generation. It paves the way for further developments and enhancements in natural language processing. Moving forward, our aim is to broaden the scope of our methodology to include the generation of more extensive texts and those tailored to specific domains. Additionally, we plan to investigate the application of evolutionary prompt techniques to a wider range of tasks involving large language models, with the goal of further pushing the frontiers of text generation technology.

\bibliographystyle{abbrv}
\bibliography{paper/paper}

\end{document}